\documentclass[11pt,a4paper]{article}
\usepackage[hyperref]{acl2020}
\usepackage{times}
\usepackage{latexsym}

\usepackage{amsmath}
\usepackage{amsfonts}
\usepackage{amssymb}
\usepackage{booktabs}
\usepackage{color} 
\usepackage{bm}
\usepackage{setspace}

\usepackage{multirow}
\usepackage{arydshln}

\usepackage{url}
\usepackage{graphicx} 
\usepackage{subfigure}

\usepackage{microtype}

\aclfinalcopy 

\title{Semantics-Aware Inferential Network for \\ Natural Language Understanding}

\author{
	Shuailiang Zhang\textsuperscript{\rm 1,2,3},
	Hai Zhao\textsuperscript{\rm 1,2,3,\thanks{Corresponding author. This paper was partially supported by National Key Research and Development Program of China (No. 2017YFB0304100) and Key Projects of National Natural Science Foundation of China (U1836222 and 61733011).}},Junru Zhou\textsuperscript{\rm 1,2,3}
	\\
	\textsuperscript{\rm 1}Department of Computer Science and Engineering, Shanghai Jiao Tong University\\
	\textsuperscript{\rm 2}Key Laboratory of Shanghai Education Commission for Intelligent Interaction\\
	and Cognitive Engineering, Shanghai Jiao Tong University, Shanghai, China\\
	\textsuperscript{\rm 3}MoE Key Lab of Artificial Intelligence, AI Institute, Shanghai Jiao Tong University, Shanghai, China\\
	{\tt zsl123@sjtu.edu.cn, zhaohai@cs.sjtu.edu.cn} \\
}
\date{}

\begin{document}
	\maketitle
	\begin{abstract} 
		For natural language understanding tasks, either machine reading comprehension or natural language inference, both semantics-aware and inference are favorable features of the concerned modeling for better understanding performance. Thus we propose a Semantics-Aware Inferential Network (SAIN) to meet such a motivation. Taking explicit contextualized semantics as a complementary input, the inferential module of SAIN enables a series of reasoning steps over semantic clues through an attention mechanism. By stringing these steps, the inferential network effectively learns to perform iterative reasoning which incorporates both explicit semantics and contextualized representations. In terms of well pre-trained language models as front-end encoder, our model achieves significant improvement on 11 tasks including machine reading comprehension and natural language inference. 	
	\end{abstract}
	
	\section{Introduction}
	Recent studies \cite{sembert,discourseaware,baidu_ernie,infoernie,paclic} have shown that introducing extra common sense knowledge or linguistic knowledge into language representations may further enhance the concerned natural language understanding (NLU) tasks that latently have a need of reasoning ability, such as natural language inference (NLI) \cite{glue,snli} and machine reading comprehension (MRC) \cite{2squad,narrativeQA}. \cite{sembert} propose incorporating explicit semantics as a well-formed linguistic knowledge by concatenating the pre-trained language model embedding with semantic role labeling embedding, and obtains significant gains on the SNLI \cite{snli} and GLUE benchmark \cite{glue}. \cite{discourseaware} use semantic information to strengthen the multi-head self-attention model, and achieves substantial improvement on NarrativeQA \cite{narrativeQA}. In this work, we propose a Semantics-Aware Inferential Network (SAIN) to refine the use of semantic structures by decomposing text into different semantic structures for compositional processing in inferential network.
	
	Questions in NLU tasks are usually not compositional, so most existing inferential networks \cite{memory,inferentialrace} input the same text at each reasoning step, which is not efficient enough to perform iterative reasoning. To overcome this problem, we use semantic role labeling to decompose the text into different semantic structures which are referred as different semantic representations of the sentence \cite{view,discourseaware}.
	
	\begin{figure}[t!]
		\centering
		\includegraphics[width=.95\linewidth]{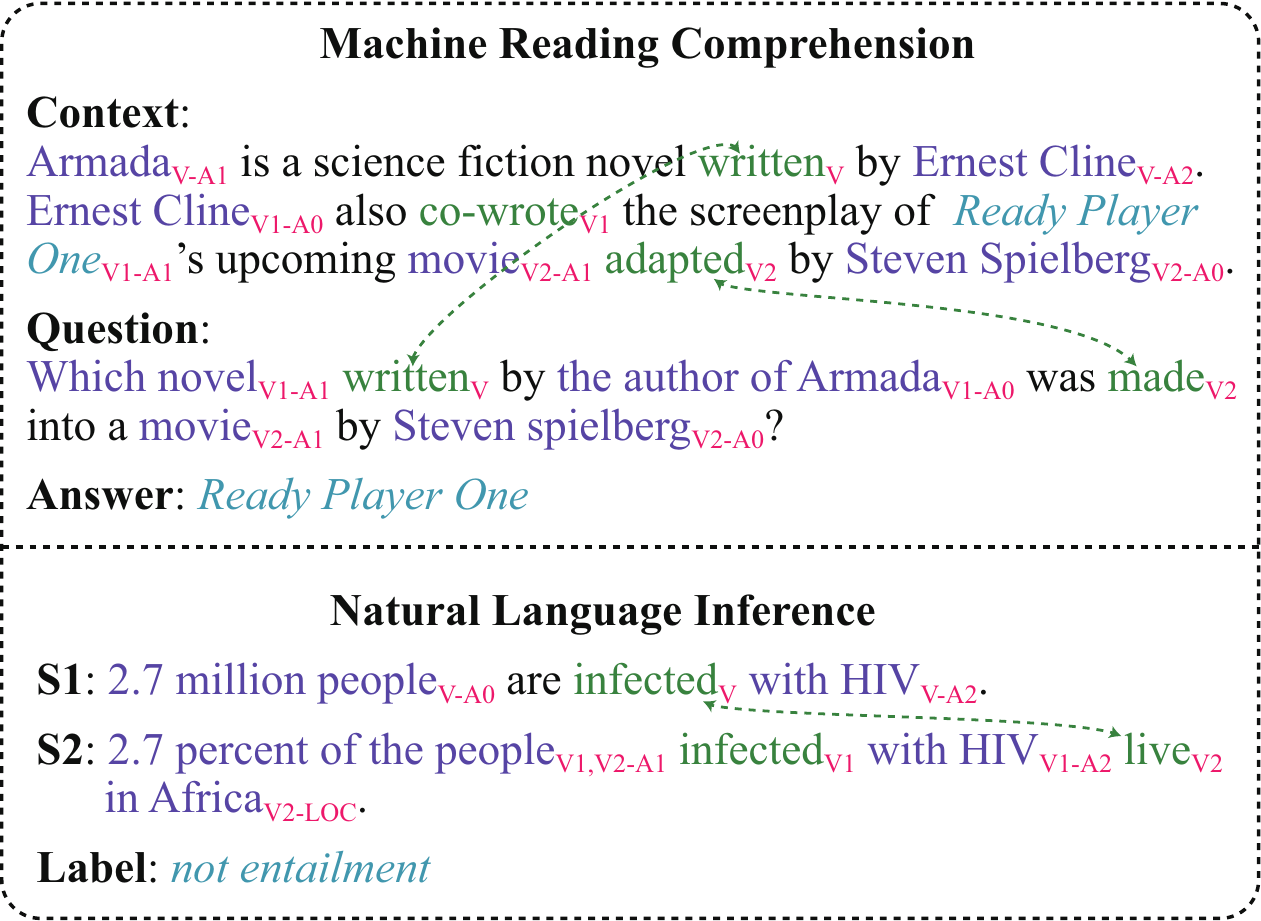}
		\caption{Examples in MRC and NLI with necessary semantic annotations. The connected predicates have important arguments to predict the answer.}
		\label{example1}
	\end{figure}

	\begin{figure*}[t!]
		\centering
		\includegraphics[width=5.4in]{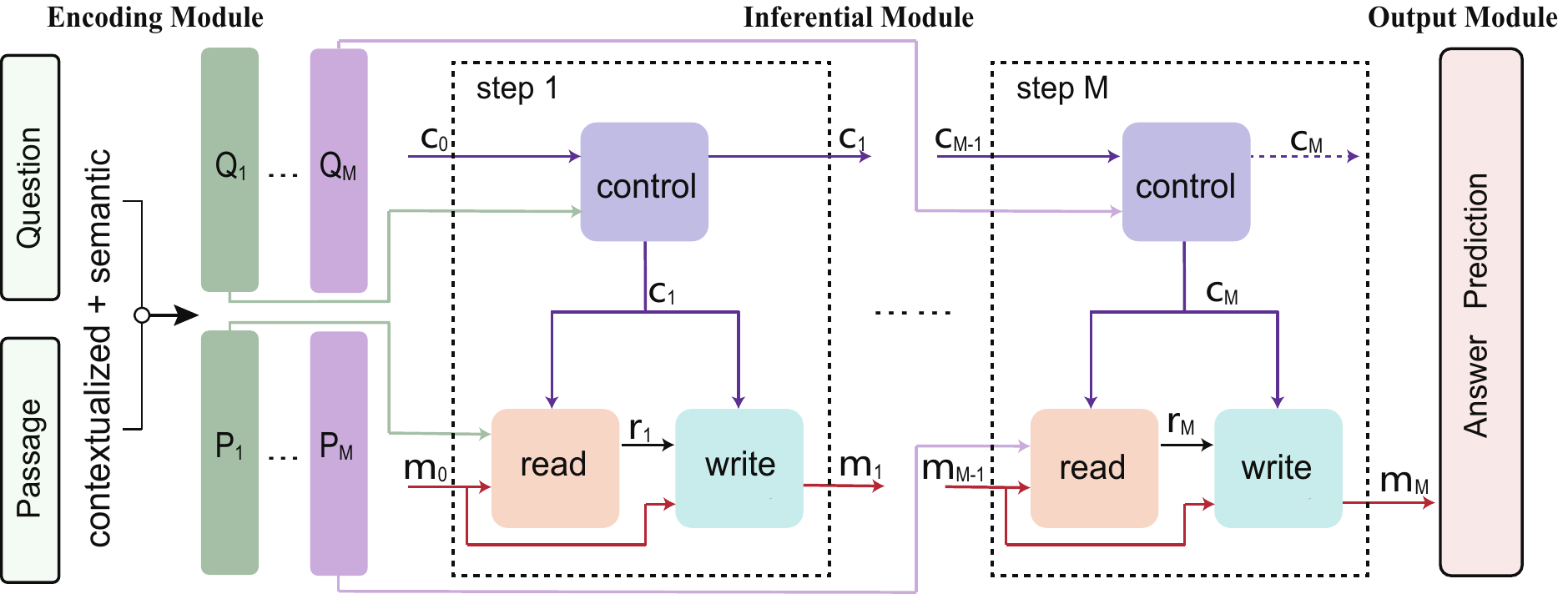}
		\caption{Overview of the framework. Here we only show the inputs and outputs of the first step and last step. The encoding module outputs $M$ semantic representations that integrate both the contextualized and semantic embedding. The model attends to $Q_i$ and $P_i$ in step $i$. The final memory state $m_{M}$ is used to predict the answer.}
		\label{figure1}
	\end{figure*}
	
	Semantic role labeling (SRL) over a sentence is to discover \textit{who did what to whom, when and why} with respect to the central meaning (usually verbs) of the sentence and present semantic relationship as predicate-argument structure, which naturally matches the requirements of MRC and NLI tasks, because questions in MRC are usually formed with \textit{who, what, how, when, why} and verbs in NLI play an important role to determine the answer. Furthermore, when there are several predicate-argument structures in one sentence, there come multiple contextual semantics. Previous neural models are usually with little consideration of modeling these multiple semantic structures which could be critical to predict the answer. 
	
	In Figure \ref{example1}, to correctly answer the MRC question, the model needs to recognize that \textit{the author of Armada} is \textit{Ernest Cline} firstly, and then knows that \textit{Ernest Cline}'s novel \textit{Ready Player One} was made into a movie by \textit{Steven spielberg}, which requires iteratively reasoning over the two predicates \textit{written} and \textit{made} because they have very similar arguments with the corresponding predicates \textit{written} and \textit{adapted} in the context. For the NLI example, if the model recognizes the predicate \textit{infected} as the central meaning in S2 and ignores the true central word \textit{live}, it probably makes wrong prediction \textit{entailment} because S1 also has a similar structure predicated on \textit{infected}. So it may be helpful to refine the use of semantic clues by integrating all the semantic information into the inference.

	We are motivated to model these semantic structures by presenting SAIN, which consists of a set of reasoning steps. In SAIN, each step attends to one predicate-argument structure and can be viewed as a cell consisting of three units: control unit, read unit and write unit, that operate over dull \emph{control} and \emph{memory} hidden states. The cells are recursively connected, where the result of the previous step acts as the context of next step. The interaction between the cells is regulated by structural constraints to perform iterative reasoning in an end-to-end way.   
	
	This work will focus on two typical NLU tasks, natural language inference (SNLI \cite{snli}, QNLI \cite{squad1}, RTE \cite{rte} and MNLI \cite{mnli}) and machine reading comprehension (SQuAD \cite{squad1,2squad} and MRQA \cite{mrqa}). Experiment results indicate that our proposed model achieves significant improvement over the strong baselines on these tasks and obtains the state-of-the-art performance on SNLI and MRQA datasets. 

	\section{Approach}
	The model framework is shown in Figure \ref{figure1}. Our model includes: 1) contextualized encoding module which obtains the joint representation of the pre-trained language model embedding and semantic embedding. 2) inferential module which consists of a set of recurrent reasoning steps/cells, where each step/cell attends to one predicate-argument structure of one sentence. 3) output module which predicts the answer based on the final memory state of the inferential module.    
	
	\subsection{Task Definition}
	For MRC task, given a passage (\textbf{P}) and a question (\textbf{Q}), the goal is to predict the answer from the given passage. For NLI task, given a pair of sentences, the goal is to judge the relationship between their meanings, such as entailment, neural and contradiction. Our model will be introduced with the background of MRC task, and the corresponding NLI implementation of our model can be regarded as a simplified case of the MRC, considering that passage and question in MRC task correspond to two sentences in NLI task.
	
	\subsection{Semantic Role Labeling}
\begin{figure}[t!]
		\centering
		\includegraphics[width=.98\linewidth]{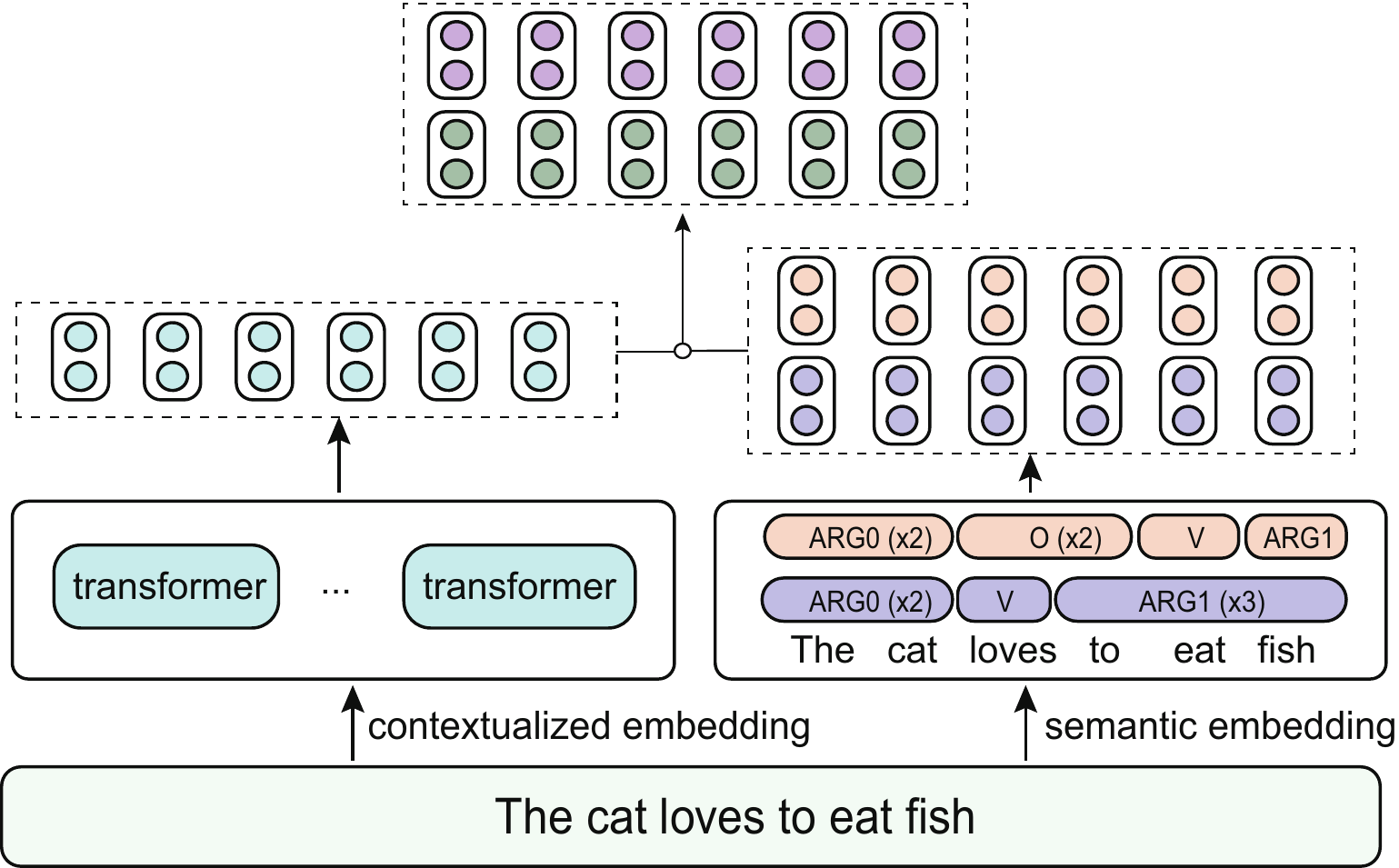}
		\caption{Different semantic representations of one sentence combined by contextualized embedding and semantic embedding.}
		\label{figuresrl}
	\end{figure}

	Semantic role labeling (SRL) is generally formulated as multi-step classification subtasks in pipeline systems to identify the semantic structures. There are a few of formal semantic frames, including FrameNet \cite{framenet} and PropBank \cite{propbank}, which generally present the semantic relationship as predicate-argument structure. When several argument-taking predicates are recognized in one sentence, we obtain multiple semantic representations of the sentence. For example, given the context sentence in Figure \ref{figuresrl} with target predicates \emph{loves} and \emph{eat}, there are two semantic structures labeled as follows,
	
	\textit{[The cat]{\tiny{ARG0}} [loves]{\tiny{V}} [to eat fish]{\tiny{ARG1}}}.
	
	\textit{[The cat]{\tiny{ARG0}} [loves to]{\tiny{O}} [eat]{\tiny{V}} [fish]{\tiny{ARG1}}}.
	
	\noindent where \text{ARG0}, \text{ARG1} represents the argument role 0, 1 of the predicate \emph{V}, respectively.
	
	\subsection{Contextual Encoding} \label{Encoding}
	\textbf{Semantic Embedding} Given the sentence $X = \{x_1,...,x_n\}$ with $n$ words and $m$ predicates ($m=2$ in Figure \ref{figuresrl}), there come $m$ corresponding labeled SRL sequences \{$L_1, L_2,..., L_m$\} with length $n$. Note this is done in data preprocessing and these labels are not updated with the following modules. These semantic role labels are mapped into vectors in dimension $d_w$ where each sequence $L_i$ is embedded as $E^{s_i} = \{e^i_1,...,e^i_n\} \in R^{n \times d_w}$.
	
	\textbf{Contextualized Embedding} With an adopted contextualized encoder, the input sequence $X = \{x_1,...,x_n\}$ is embedded as $E^w = \{e_1,...,e_{n_s}\} \in R^{n_s \times d_s}$, where $d_s$ is hidden state size of the encoder and $n_s$ is the tokenized sequence length. 
	
	\textbf{Joint embedding} Note that the input sequence may be tokenized into subwords. Then the tokenized sequence of length $n_s$ is usually longer than the SRL sequence of length $n$. To align these two sequences, we extend the SRL sequence to length $n_s$ by assigning the subwords the same label with original word\footnote{For example, if $x_j$ is tokenized into three subwords \{$x_{j_1}, x_{j_2}, x_{j_3}$\}, then $E^{s_i} = \{e^i_1,...,e^i_j,..., e^i_n\}$ is extended to $E^{s_i} = \{e^i_1,..., e^i_j,e^i_j,e^i_j,..., e^i_n\}$}. The aligned contextualized and semantic embeddings are then concatenated as the joint embedding\footnote{We also tried summation and multiplication, but experiments show that concatenation is the best.} for the sequence $E^{X_i} = [E^{s_i};E^w] \in R^{n_s \times d}$, where $d=d_s+d_w$. 
	
	Different sentences have various numbers of predicate-argument structures, here we set the maximum number as $M$ for ease of calculation\footnote{The sentences without enough number of semantic structures are padded to $M$ structures where all the labels are assigned to \emph{O}. For example, the sentence in Figure \ref{figuresrl} is padded as [The cat loves to eat fish]{\tiny{O}}.}. So for MRC, the passage and question both have $M$ encoded representations where $E^{P} =\{E^{P_1},...,E^{P_M}\} \in R^{M \times |P| \times d}$ and $E^{Q} = \{E^{Q_1},...,E^{Q_M}\}\in R^{M \times |Q| \times d}$, where $|P|$, $|Q|$ are the subwords numbers of passage and question.

	\subsection{Inferential Network} 
	The inferential module performs explicit multi-step reasoning by stringing together $M$ cells, where each attends to one semantic structure of the sentence. Each cell has three operation units: control unit, read unit and write unit, iteratively aggregating information from different semantic structures.
	
	For MRC, each reasoning step attends to one semantic structure of each sentence from passage and question, respectively. So passage $E^{P_i} = \{p_{i,1},...,p_{i,|P|}\}$ and question $E^{Q_i} = \{q_{i,1},...,q_{i,|Q|}\}$ are the input sequences for step $i$. Besides, we use biLSTM to get the overall question representation $bq_i = [\overrightarrow{q_{i,1}};\overleftarrow{q_{i,|Q|}}]\in R^{2d}$.  	 
	
	\textbf{Reasoning Cell} The reasoning cell is a recurrent cell designed to capture information from different semantic structures. For each step $i = 1,...,M$ in the reasoning process, the $i^{th}$ cell maintains two hidden states: \textbf{control} $\bm{c_i}$ and \textbf{memory $\bm{m_i}$}, with dimension $d$. The control $\bm{c_i}$ retrieves information from $E^{Q_i}$ by calculating a soft attention-based weighted average of the question words. The memory $\bm{m_i}$ holds the intermediate results from the reasoning process up to the $i^{th}$ step by integrating the preceding hidden state $\bm{m_{i-1}}$ with the new information $\bm{r_i}$ retrieved from the passage $E^{P_i}$. 

	\begin{figure}[t!]
		\centering
		\includegraphics[width=.98\linewidth]{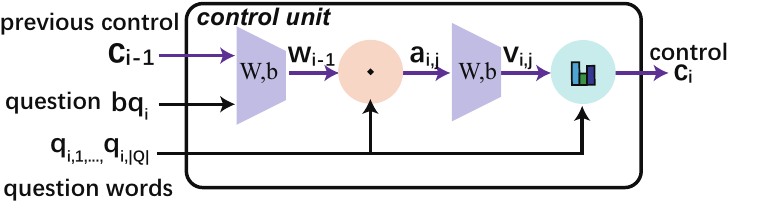}
		\caption{The control unit.}
		\label{control}
	\end{figure}
		
	There are three units in each cell: control unit, read unit and write unit, which work together to perform iterative reasoning. The control unit retrieves the information from the question, updating the control hidden state $\bm{c_i}$. The read unit extracts relevant information from the passage and outputs extracted information $\bm{r_i}$. The write unit integrates $\bm{c_i}$ and $\bm{r_i}$ into the memory $\bm{m_{i-1}}$, producing a new memory $\bm{m_i}$. In the following, we give the details of these three units. All the vectors are of dimension $d$ unless otherwise stated.    
	
	The \textbf{control unit} (Figure \ref{control}) attends to  the $i^{th}$ semantic structure of the question $E^{Q_i}$ at step $i$ and updates the control state $\bm{c_i}$ accordingly. Firstly, we combines the overall question representation $\bm{bq_i}$ and preceding reasoning operation $\bm{c_{i-1}}$ into $\bm{w_i}$ through a linear layer. Subsequently, we calculate the similarity between $\bm{w_i}$ and each question word $\bm{q_{i,j}}$, and pass the result through a softmax layer, yielding an attention distribution over the question words. Finally, we sum the words over this distribution to get the new control $\bm{c_i}$. The calculation details are as follows:
	
	\begin{equation} \nonumber
	\begin{split}
	w_i &= W^{d \times 2d}[c_{i-1}, bq_i] + b^d \\
	a_{i,j} &= W^{1 \times d}(w_i \odot q_{i,j}) + b^1 \\
	v_{i,j} &= \textit{Softmax}(a_{i,j}), j = 1,...,|Q| \\
	c_i &= \sum_{j=1}^{|Q|} v_{i,j} \cdot q_{i,j}
	\end{split}
	\end{equation} 
	where $W^{d \times 2d}$, $W^{1 \times d}$, $b^d$ and $b^1$ are learnable parameters, $|Q|$ is the subwords numbers of question.

\begin{figure}[t]
		\centering
		\includegraphics[width=\linewidth]{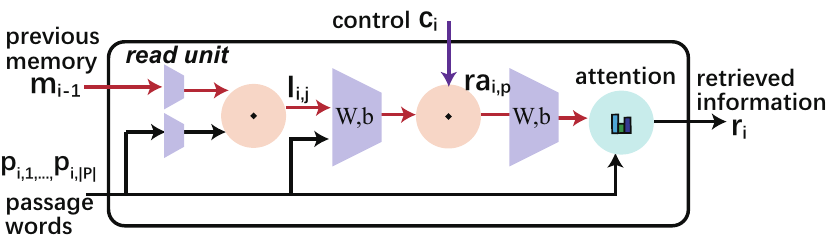}
		\caption{The read unit.}
		\label{read}
	\end{figure}

	The \textbf{read unit} (Figure \ref{read}) inspects the $i^{th}$ semantic structure of the passage $E^{P_i}$ at step $i$ and retrieves the information $\bm{r_i}$ to update the memory. Firstly, we compute the interaction between every passage word $p_{i,p}$ and the memory $\bm{m_{i-1}}$, resulting in $I_{i,p}$ which measures the relevance of the passage word to the preceding memory. Then, $I_{i,p}$ and $p_{i,p}$ are concatenated and passed through a linear transformation, yielding $\hat{I}_{i,p}$ which considers both the new information from $E^{P_i}$ and the information related to the prior intermediate result. Finally, aiming to retrieve the information relevant to the question, we measure the similarity between $\hat{I}_{i,p}$ and $c_i$ and pass the result through a softmax layer which produces an attention distribution over the passage words. This distribution is used to get the weighted average $\bm{r_i}$ over the passage. The calculation is detailed as follows: 
	\begin{equation} \nonumber
	\begin{split}
	I_{i,p} &= [W_1^{d \times d}m_{i-1} + b_1^d] \odot [W_2^{d \times d}p_{i,p} + b_2^d] \\
	\hat{I}_{i,p} &= W^{d \times 2d}[I_{i,p}, p_{i,p}] + b^d \\
	ra_{i,p} &= W^{d \times d}(c_i \odot \hat{I}_{i,p}) + b^d \\
	rv_{i,p} &= \textit{Softmax}(ra_{i,p}), p = 1,...,|P| \\
	r_i &= \sum_{p=1}^{|P|} rv_{i,p} \cdot p_{i,p}
	\end{split}
	\end{equation}
	where all the $W$ and $b$ are learnable parameters, $|P|$ is the subwords numbers of passage.
	
	\begin{figure}[b]
		\centering
		\includegraphics[width=\linewidth]{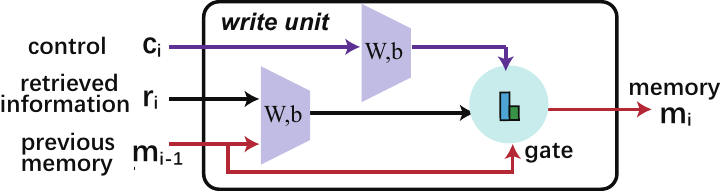}
		\caption{The write unit.}
		\label{write}
	\end{figure}
	
	The \textbf{write unit} (Figure \ref{write}) is responsible for integrating the information retrieved from the read unit $\bm{r_i}$ with the preceding memory $\bm{m_{i-1}}$, guided by the $i^{th}$ reasoning operation $\bm{c_i}$ from the question. Specificly, a sigmoid gate is used when combining the previous memory state $\bm{m_{i-1}}$ and the new memory candidate $\bm{m_i^r}$. The calculation details are as follows: 
	\begin{equation} \nonumber
	\begin{split}
	m_i^r &= W^{d \times 2d}[r_i, m_{i-1}] + b^d \\
	\hat{c}_i &= W^{1 \times d} c_i + b^1 \\
	m_i &= \sigma(\hat{c}_i) m_{i-1} + (1 - \sigma(\hat{c}_i)) m_i^r 
	\end{split}
	\end{equation}

	\subsection{Output Module}
	For MRC, the output module predicts the final answer to the question based on the set of memory states \{$m_\text{1}$,...,$m_\text{M}$\} produced by the inferential module. For MRC, we calculate the similarity between the $i^{th}$ memory $m_\text{i} \in R^d$ and each passage word $p_{i,p}$ in $i^{th}$ semantic passage representation $E^{P_i}$, resulting in $\hat{E}^{P_i}$, $i = \text{1},...,\text{M}$. We concatenate $\hat{E}^{P_1},...,\hat{E}^{P_M}$ as the final passage representation $\hat{E}^P \in R^{|P| \times Md}$ which is then passed to a linear layer to get the start and end probability distribution $p_s$, $p_e$ on each position. Finally, a cross entropy loss is computed:
	\begin{equation} \nonumber
	\begin{split}
	mp_{i,p} &= m_i \cdot p_{i,p} \\
	\hat{E}^{P_i} &= [mp_{i,1},...,mp_{i,|P|}] \in R^{|P| \times d} \\
	E &= [\hat{E}^{P_1},...,\hat{E}^{P_M}] \in R^{|P| \times Md} \\
	[p_s, p_e] &= E W^{Md \times 2} \in R^{|P| \times 2}\\
	Loss &= \frac{1}{2} \textit{CE}(p_s, y_s) + \frac{1}{2} \textit{CE}(p_e, y_e) 
	\end{split}
	\end{equation} 
	where $y_s$ and $y_e$ are the true start and end probability distribution. $p_s$, $p_e$, $y_s$ and $y_e$ are all with size $R^{|P|}$. $CE(\cdot)$ indicates the cross entropy function. 
	
	For NLI, the final memory state $m_\text{M}$ is directly passed to a linear layer to produce the probability distribution over the labels: $p = m_\text{M} W^{d \times N} \in R^N$. Cross entropy is used as the metric: $\text{Loss} = \textit{CE}(p, y)$, where $N$ is the number of labels. $p\in R^N$ is the predicted probability distribution over the labels and $y\in R^N$ is the true label distribution.
	
	\section{Experiments}
	\subsection{Data and Task Description}
	\textbf{Machine Reading Comprehension} We evaluate our model on extractive MRC such as SQuAD \cite{2squad} and MRQA\footnote{https://github.com/mrqa/MRQA-Shared-Task-2019.} \cite{mrqa} where the answer is a span of the passage. MRQA is a collection of existing question-answering related MRC datasets, such as SearchQA \cite{searchQA}, NewsQA \cite{newsQA}, NaturalQuestions \cite{naturalQuestions}, TriviaQA \cite{TriviaQA}, etc. All these datasets as shown in Table \ref{smrqa} are transformed into SQuAD style where given the passage and question, the answer is a span of the passage.

	\textbf{Natural Language Inference} Given a pair of sentences, the target of natural language inference is to judge the relationship between their meanings, such as entailment, neural and contradiction. We evaluate on 4 diverse datasets, including Stanford Natural Language Inference (SNLI) \cite{snli}, Multi-Genre Natural Language Inference (MNLI) \cite{mnli}, Question Natural Language Inference (QNLI) \cite{squad1} and Recognizing Textual Entailment (RTE) \cite{rte}.
	\begin{table}[t]
		\begin{center}
			\begin{tabular}{lllll}
				\toprule
				Dataset &\#train &\#dev &$|\text{P}|$&$|\text{Q}|$ \\
				\hline
				NewsQA &74,160 &4,211 &599 &8\\
				TriviaQA &61,688 &7,785 &784 &16\\
				SearchQA &117,384 &16,980 &749 &17\\
				HotpotQA  &72,928 &5,904 &232 &22\\
				NaturalQA  &104,071 &12,836 &153 &9\\
				\bottomrule
			\end{tabular} 
		\end{center}
		\caption{\label{smrqa} Statistics of MRQA datasets. \#train and \#dev are the number of examples in train and dev set. $|\cdot|$ denotes the average length in tokens.}
	\end{table}
	
	\begin{table*}[t]
		\begin{center}
			\resizebox{.99\linewidth}{!}{
			\begin{tabular}{lccccc|l}
				\toprule
				& NewsQA &TriviaQA &SearchQA &HotpotQA &NaturalQA &\textbf{(Avg.)} \\
				\hline
				MTL$_\text{base}$ \cite{mrqa}     &66.8 &71.6  &76.7 &76.6 &77.4 &73.8\\
				MTL$_\text{large}$ \cite{mrqa}   &66.3 &74.7  &79.0 &79.0 &79.8 &75.8\\
				CLER \cite{cler} &69.4 &75.6 &79.0 &79.8 &79.8 &76.7\\
				BERT$_\text{large}$ \cite{spanbert}  &68.8 &77.5  &81.7 &78.3 &79.9 &77.3\\
				HLTC \cite{hltc}  &72.4 &76.2 &79.3 &80.1 &80.6 &77.7 \\
				SemBERT$^*$ \cite{sembert}  &69.1 &78.6  &82.4 &78.6 &80.3 &77.8\\
				SpanBERT \cite{spanbert} &73.6 &83.6 &84.8 &83.0 &82.5 &81.5\\
				\hdashline
				BERT$^*$$_\text{base}$     &66.2 &71.5 &77.0 &75.0 &77.5 &73.4\\
				BERT$^*$$_\text{large}$     &69.2 &77.4 &81.5 &78.2 &79.4 &77.2\\
				SpanBERT$^*$     &73.0 &83.1 &83.5 &82.5 &81.9 &80.8\\
				\hline
				Our Models &&&&&&\\
				SAIN\textsubscript{BERT$_\text{base}$} &67.9 &72.3 &77.8  &77.4 &78.6 &74.8\\
				SAIN\textsubscript{BERT$_\text{large}$} &72.1 &80.1 &83.4 &79.4 &82.0 &79.4\\
				SAIN\textsubscript{SpanBERT} &74.2 &84.5 &84.4 &83.4 &82.7 &\textbf{81.9}\\
				\bottomrule
			\end{tabular}}
		\end{center}
		\caption{\label{tablemrqa} Performance (F1) on five MRQA tasks. Results with $^*$ are our implementations. \textbf{Avg} indicates the average score of these datasets. All these results are from single models.}
	\end{table*}
	
	\begin{table*}[t]
		\begin{center}
			\begin{tabular}{lcccccc|cccc}
				\toprule
				\bf Model         &\multicolumn{2}{c}{MNLI-m/mm} &QNLI &RTE &SNLI &\textbf{(Avg.)} &\multicolumn{2}{c}{SQuAD 1.1} &\multicolumn{2}{c}{SQuAD 2.0}\\
				&Acc &Acc &Acc &Acc&Acc& &EM &F1 &EM &F1\\
				\hline
				BERT$_\text{base}$    &84.6&83.4 &89.3 &66.4 &90.7&82.9 &80.8 &88.5 &77.1$^*$ &80.3$^*$\\
				BERT$_\text{large}$  &86.7&85.9 &92.7 &70.1 &91.1 &85.3&84.1 &90.9  &80.0 &83.3 \\
				SemBERT$_\text{base}$    &84.4&84.0 &90.9 &69.3$\dagger$ &91.0$^*$ & 83.9&\_&\_&\_&\_\\
				SemBERT$_\text{large}$  &87.6 &86.3 &\bf{94.6} &70.9$\dagger$ &91.6&86.2  &84.5$^*$ &91.3$^*$ &80.9 &83.6 \\
				
				\hline
				Our Models&&&&&&&&&&\\
				SAIN\textsubscript{BERT$_\text{base}$} &84.9&85.0 &92.1 &72.0  &91.2&85.1 &82.2 &89.3  &79.4 &82.0 \\
				SAIN\textsubscript{BERT$_\text{large}$} &\bf{87.7} &\bf{87.3} &94.5 &\bf{73.9}  &\bf{91.7} &\bf{87.1} &\bf{85.4} &\bf{91.9}  &\bf{82.8} &\bf{85.4} \\
				\bottomrule
			\end{tabular}
		\end{center}
		\caption{\label{tableall} Experiment results on MNLI, QNLI, RTE, SNLI and SQuAD 1.1, SQuAD 2.0. The results of BERT and SemBERT are from \cite{Devlin-18} and \cite{sembert}. $\dagger$ indicates the results of SemBERT without random restarts and distillation. Results of SQuAD are tested on development sets. Results with $^*$ are our implementations. \textbf{Avg} indicates the average score of these datasets. All these results are from single models.}
	\end{table*}
	
	\subsection{Implementation Details}
	To obtain the semantic role labels, we use the SRL system of \cite{hesrl} as implemented in AllenNLP \cite{allen} that splits sentences into tokens and predicts SRL tags such as \emph{ARG0}, \emph{ARG1} for each verb. We use \emph{O} for non-argument words and \emph{V} for predicates. The dimension of SRL embedding is set to 30 and performance does not change significantly when setting this number to 10, 50 or 100. The maximum number of predicate-argument structures (reasoning steps) $M$ is set to 3 or 4 for different tasks. 
	
	Our model framework is based on the Pytorch implementation of transformers\footnote{https://github.com/huggingface/transformers.}. We use Adam as our optimizer with initial learning rate 1e-5 and warm-up rate of 0.1. The batch size is selected in \{8, 16, 32\} with respect to the task. The total parameters vary from 355M (total steps $M=1$) to 362M ($M=7$), increasing 20M to 27M parameters compared to BERT (335M).
	
	\subsection{Overall Results}
	Our main comparison models are the BERT baselines (BERT \cite{Devlin-18} and SpanBERT \cite{spanbert}) and SemBERT \cite{sembert}. SemBERT improves the language representation by concatenating the BERT embedding and semantic embedding, where embeddings from different predicate-argument structures are simply fused as one semantic representation by using one linear layer. We compare our model to these baselines on 11 benchmarks including 5 MRQA datasets, 4 NLI tasks and 2 SQuAD datasets in Tables \ref{tablemrqa} and \ref{tableall}.
	
	\textbf{SAIN vs. BERT/SpanBERT baselines} Compared to BERT \cite{Devlin-18}, our model achieves general improvements, including 2.1\% (79.4 vs. 77.3), 1.8\% (87.1 vs. 85.3), 1.6\% (88.7\% vs. 87.1 \%) average improvement on 5 MRQA, 4 NLI and 2 SQuAD datasets. Our model also outperforms other BERT based models CLER \cite{cler} and HLTC \cite{hltc} on MRQA. We also compare with SpanBERT \cite{spanbert} on MRQA datasets and our model outperforms this baseline by 0.4\% (81.9 vs. 81.5) in average F1 score. To the best of our knowledge, we achieve state-of-the-art performance on MRQA (dev sets) and SNLI.
	
	\begin{table}[t!]
		\begin{center}
			\resizebox{.96\linewidth}{!}{
				\begin{tabular}{llll}
					\toprule
					&RTE &SQuAD 1.1 &SQuAD 2.0 \\
					\hline
					SAIN &\bf{73.4} 　&\bf{91.9} &\bf{85.4} \\
					\hline
					w/o IM &71.2 (-2.3) &90.6 (-1.3) &82.5 (-1.9) \\
					w/o SI &71.5 (-1.9) &90.1 (-1.8) &83.1 (-2.3) \\
					w/o IR &72.0 (-1.4)　&90.9 (-1.0) &83.2 (-2.2) \\
					\bottomrule
			\end{tabular}}
		\end{center}
		\caption{\label{ablation} Ablation study on RTE, SQuAD 1.1 and SQuAD 2.0 (F1). We use BERT$_\text{large}$ as contextual encoder here. The definition of IM, SI and IR is detailed in Section \ref{anno}.}
	\end{table}

	\textbf{SAIN vs. SemBERT}  Our SAIN outperforms SemBERT on all tasks, including 1.6\% (79.4 vs. 77.8), 0.9\% (87.1 vs. 86.2) and 1.3\% (86.4 vs. 85.1) average improvement on MRQA, NLI and SQuAD datasets. We attribute the superiority of our SAIN to its more refined use of semantic clues in terms of inferential network rather than SemBERT which simply encodes all predicate-argument structures into one embedding.
	
	\subsection{Ablation Study} \label{anno}
	To evaluate the contribution of key components in our model, we perform ablation studies on the RTE and SQuAD 2.0 dev sets as shown in Table \ref{ablation}. Here we focus on these components: (1) the whole inferential module (IM); (2) the semantic information (SI); (3) iterative reasoning (IR) that different reasoning cells attend to different predicate-argument structures. To evaluate their contribution, we perform experiments respectively by: (1) IM: removing the inferential module and simply combining the BERT embedding with semantic embeddings from different predicate-argument structures; (2) SI: removing all the semantic embeddings; (3) IR: combining all semantic embeddings from different predicate-argument structures as one and every reasoning step taking the same semantic embedding.

	\begin{figure}[t]
		\centering
		\includegraphics[width=.95\linewidth]{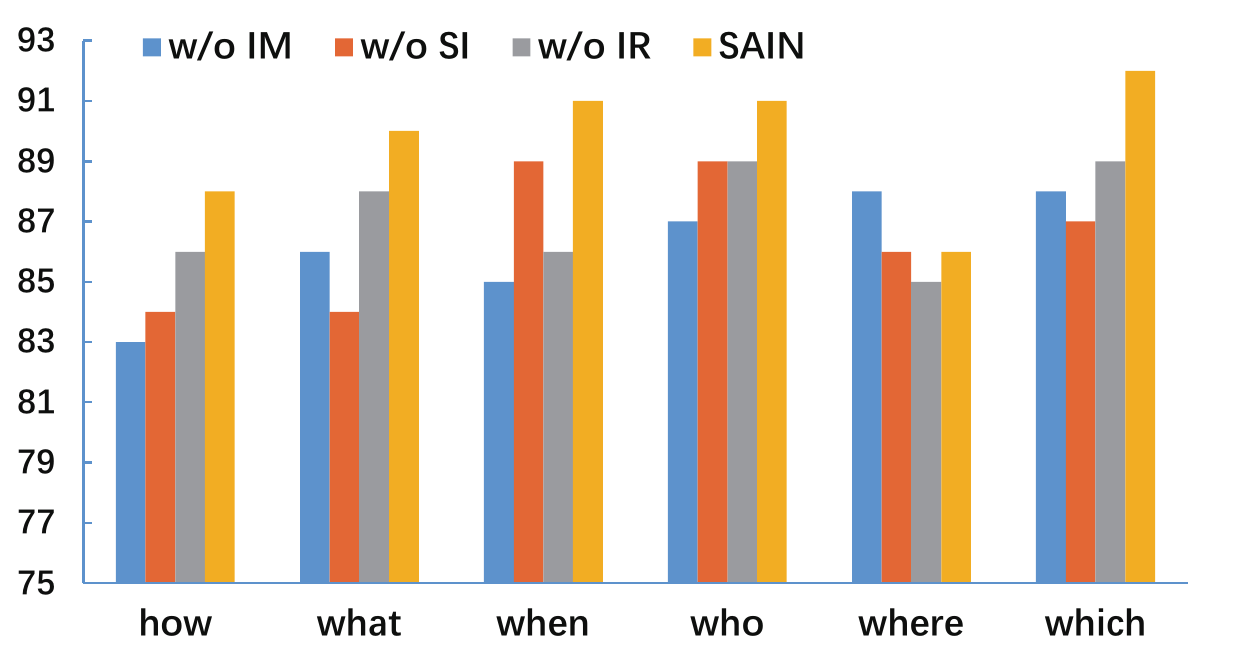}
		
		\caption{Performance on different question types, tested on the SQuAD 1.1 development set. BERT$_{\text{base}}$ is used as contextual encoder here. The definition of IM, SI and IR is detailed in Section \ref{anno}.}
		\label{figuretype}
	\end{figure}

	
	As displayed in Table \ref{ablation}, the ablation on all evaluated components results in performance drop which indicates that all the key components (the inferential module, the semantic information and iterative reasoning process) are indispensable for the model. Particularly, the ablation on iterative reasoning proves that it is necessarily helpful that the model attends to different predicate-argument structures in different reasoning steps.
	
	Furthermore, Figure \ref{figuretype} shows the ablation results on different question types, tested on sampled examples from SQuAD 1.1. The full SAIN model outperforms all other ablation models on all question types except the \textit{where} type questions, which again proves that integrating the semantic information of (\textit{who did what to whom, when} and \textit{why}) contributes to boosting the performance of MRC tasks where questions are usually formed with \textit{who, what, how, when} and \textit{why}.
	
	\subsection{Influence of Semantic Information}
	
\begin{figure}[t!]
		\centering
		\includegraphics[width=.95\linewidth]{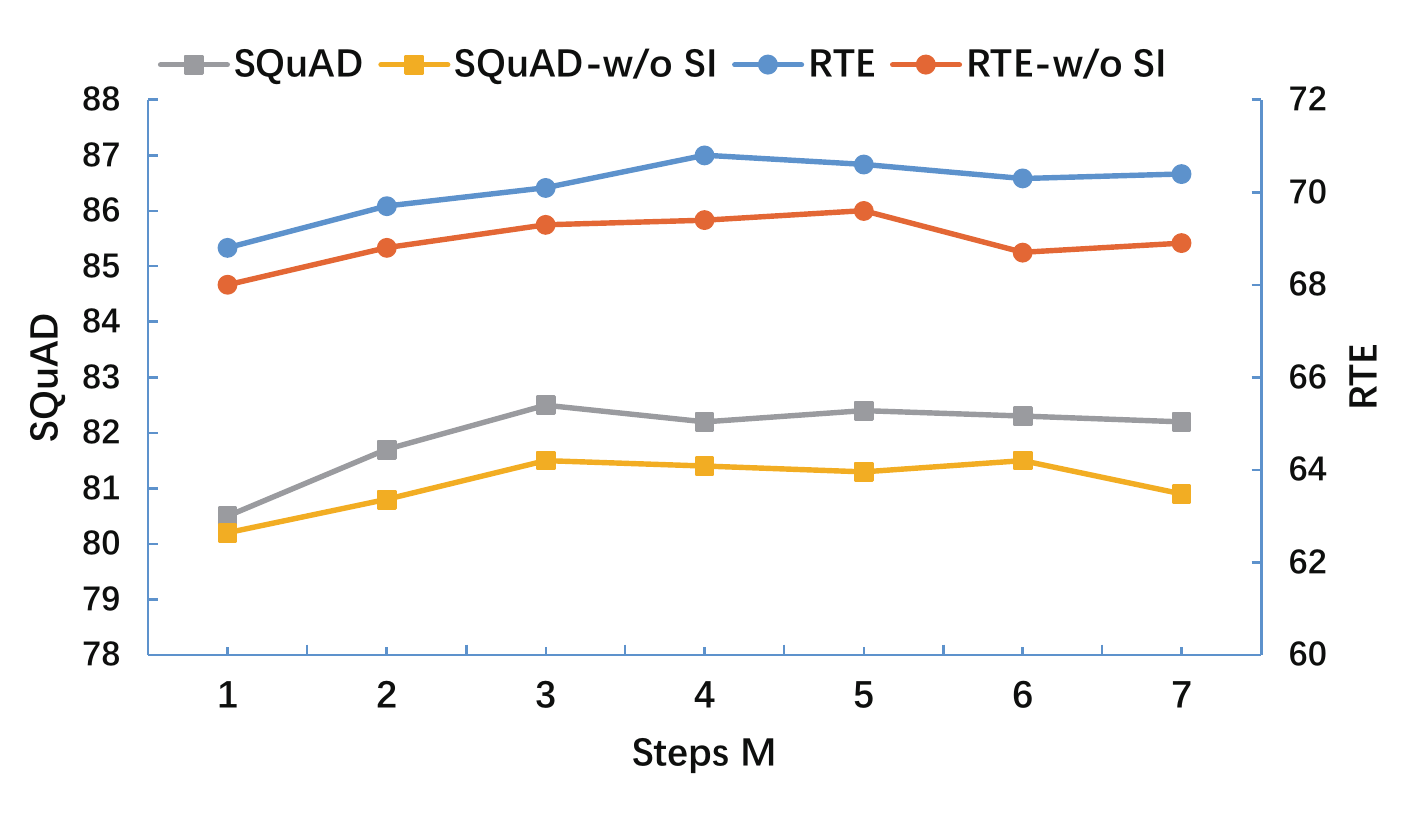}
		\caption{Results on the dev sets of SQuAD 2.0 and RTE when selecting different reasoning steps $M$. We use BERT$_\text{base}$ as contextual encoder here. SQuAD/RTE-w/o SI indicates the results without using any semantic information.}
		\label{setp}
	\end{figure}

	To further investigate the influence of semantic information, Figure \ref{setp} shows the performance comparison of whether to use the semantic information with different numbers of reasoning steps $M$ (from 1 to 7). The highest performance is achieved when $M$ is set to 3 on SQuAD, 4 on RTE. The results indicate that semantic information consistently contributes to the performance increase, although the inferential network is strong enough. 
	
	To investigate influence of the accuracy of the labeler, we randomly tune specific proportion [0, 20\%, 40\%] of labels into random error ones. The scores of SQuAD 2.0 and RTE are respectively [85.4, 83.2, 82.6] and [73.4, 71.8, 71.2], which indicate that the model benefits from high-accuracy labeler but can still maintain the performance even using some noisy labels.

	\subsection{Influence of Inferential Mechanism}
\begin{figure}[t!]
		\includegraphics[width=.98\linewidth]{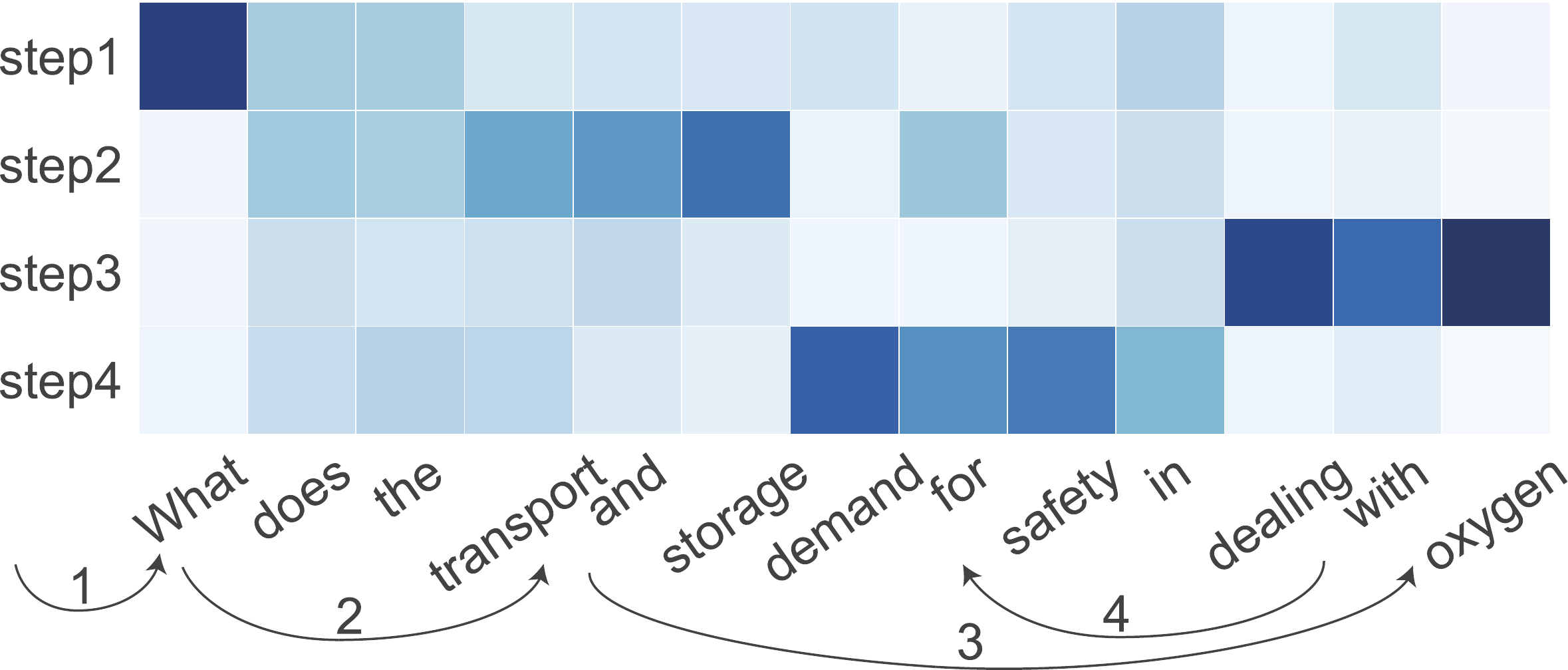}
		\caption{Transformation of attention distribution at each reasoning step, showing how the model iteratively retrieves information from the question.}
		\label{heatmap}
	\end{figure}

	To obtain better insight into the underlying reasoning processes, we study the visualization of the attention distributions during the iterative computation, and provide examples in Table \ref{casestudy} and Figure \ref{heatmap}. Table \ref{casestudy} shows a relatively complex question that is correctly answered by our model, but wrongly predicted by SemBERT \cite{sembert}. In this example, there is misleading contextual similarity between words ``\textit{store and transmit}" in sentence S1 and ``\textit{transport and storage}" in the question which may lead the model to wrong answer in S1, such as ``\textit{fuel}" by SemBERT. To overcome this misleading, the model needs to recognize the central connection predicates ``\textit{demand}" and ``\textit{requires}" between the question and passage, then extract the correct answer ``\textit{special training}" in S2.
	
	Figure \ref{heatmap} shows how our model retrieves information from different semantic structures of the question in each reasoning step. The model first focuses on the word ``\textit{what}", working to retrieve a noun. Then it focuses on the arguments ``\textit{transport}" and ``\textit{storage}" in step 2 but gets around these words in step 3, and attends to the second verb phrase ``\textit{dealing with oxygen}", taking the model's attention away from sentence S1. Finally, the model focuses on the main meaning of the question: ``\textit{demand for security}" and predicts the correct answer ``\textit{special training}" in sentence S2, with respect to the semantic similarity between words ``\textit{demand for safety}" and ``\textit{requires to ensure}". This example intuitively explains why our model benefits from the iterative reasoning where each step only attends to one semantic representation.

	\begin{table}[t!] 
		\begin{center}
			\resizebox{.96\linewidth}{!}{
				\begin{tabular}{p{7.3cm}}
					\toprule
					\textbf{Passage}: (S1) \textit{Steel pipes and storage vessels used to store and transmit both gaseous and liquid oxygen will act as a \textcolor[RGB]{168,0,0}{fuel};} (S2) \textit{and therefore the design and manufacture of oxygen systems {requires} \textcolor[RGB]{0,128,0}{special training} to ensure that ignition sources are minimized.}  \\ 
					\midrule
					\textbf{Question}: \textit{What does the transport and storage {demand} for safety in dealing with oxygen?}\\ 
					\textbf{Golden Answer}: \textcolor[RGB]{0,128,0}{\textit{special training}} \\
					\hline   
					\textbf{SemBERT}: \textcolor[RGB]{168,0,0}{\textit{fuel}} \quad  \textbf{SAIN}: \textcolor[RGB]{0,128,0}{\textit{special training}} \\
					\bottomrule
			\end{tabular}}
		\end{center}
		\caption{\label{casestudy} One example that is correctly predicted by SAIN, but wrongly predicted by SemBERT.}
	\end{table}

	\section{Related Work} 
	\textbf{Semantic Information for MRC}        
	Using semantic information to enhance the question answering system is one effective method to boost the performance. \cite{semanticstructure} first stress the importance of semantic roles in dealing with complex questions. \cite{semanticrole} introduce a general framework for answer extraction which exploits semantic role annotations in the FrameNet \cite{framenet} paradigm. \cite{lexical} propose to solve the answer selection problem using enhanced lexical semantic models. More recently, \cite{sembert} propose to strengthen the language model representation by fusing explicit contextualized semantics. \cite{discourseaware} apply linguistic annotations to a discourse-aware semantic self-attention encoder which is employed for reading comprehension on narrative texts. In this work, we propose to use inferential model to recurrently retrieve different predicate-argument structures, which presents a more refined way using semantic clues and thus is essentially different from all previous methods. 
	
	\textbf{Inferential Network} 
	To support inference in neural network, exiting models either rely on structured rule-based matching methods \cite{sun-reading} or multi-layer memory networks \cite{memory,gatedmemory}, which either lack end-to-end design or no prior structure to subtly guide the reasoning direction.
	
	Another related works are on Visual QA, aiming to answer the question with regards to the given image. In particular, \cite{relation} propose a relation net but restricted the relational question such as comparison. Later, for compositional question, \cite{mac} introduce an iterative network that separates memory and control to improve interpretability. Our work leverages such separate design, dedicating to inferential NLI and MRC tasks, where the questions are usually not compositional. 
	
	To overcome the difficulty of applying inferential network into general NLU tasks, and passingly refine the use of multiple semantic structures, we propose SAIN which naturally decomposes text into different semantic structures for compositional processing in inferential network.
	
	\section{Conclusion}
	This work focuses on two typical NLU tasks, machine reading comprehension and natural language inference by refining the use of semantic clues and inferential model. The proposed semantics-aware inferential network (SAIN) is capable of taking multiple semantic structures as input of an inferential network by closely integrating semantics and reasoning steps in a creative way. Experiment results on 11 benchmarks, including 4 NLI tasks and 7 MRC tasks, show that our model outperforms all previous strong baselines, which consistently indicate the general effectiveness of our model\footnote{Our model can be easily adapted to other language models such as ALBERT, which is left for future work.}.

\bibliography{acl2020}

\begin{thebibliography}{35}
\expandafter\ifx\csname natexlab\endcsname\relax\def\natexlab#1{#1}\fi

\bibitem[{Baker et~al.(1998)Baker, Fillmore, and Lowe}]{framenet}
Collin~F. Baker, Charles~J. Fillmore, and John~B. Lowe. 1998.
\newblock \href {https://doi.org/10.3115/980845.980860} {The {B}erkeley
  {F}rame{N}et project}.
\newblock In \emph{36th Annual Meeting of the Association for Computational
  Linguistics and 17th International Conference on Computational Linguistics
  (ACL)}, pages 86--90.

\bibitem[{Bentivogli et~al.(2009)Bentivogli, Dagan, Dang, Giampiccolo, and
  Magnini}]{rte}
Luisa Bentivogli, Ido Dagan, Hoa~Trang Dang, Danilo Giampiccolo, and Bernardo
  Magnini. 2009.
\newblock {The Fifth PASCAL Recognizing Textual Entailment Challenge}.
\newblock In \emph{In Proc Text Analysis Conference (TAC’09)}.

\bibitem[{Bowman et~al.(2015)Bowman, Angeli, Potts, and Manning}]{snli}
Samuel~R. Bowman, Gabor Angeli, Christopher Potts, and Christopher~D. Manning.
  2015.
\newblock \href {https://doi.org/10.18653/v1/D15-1075} {A large annotated
  corpus for learning natural language inference}.
\newblock In \emph{Proceedings of the 2015 Conference on Empirical Methods in
  Natural Language Processing (EMNLP)}, pages 632--642.

\bibitem[{Devlin et~al.(2019)Devlin, Chang, Lee, and Toutanova}]{Devlin-18}
Jacob Devlin, Ming-Wei Chang, Kenton Lee, and Kristina Toutanova. 2019.
\newblock \href {https://doi.org/10.18653/v1/N19-1423} {{BERT: Pre-training of
  Deep Bidirectional Transformers for Language Understanding}}.
\newblock In \emph{Proceedings of the 2018 Conference of the North {A}merican
  Chapter of the Association for Computational Linguistics: Human Language
  Technologies (NAACL)}, pages 4171--4186.

\bibitem[{Dunn et~al.(2017)Dunn, Sagun, Higgins, G{\"{u}}ney, Cirik, and
  Cho}]{searchQA}
Matthew Dunn, Levent Sagun, Mike Higgins, V.~Ugur G{\"{u}}ney, Volkan Cirik,
  and Kyunghyun Cho. 2017.
\newblock \href {http://arxiv.org/abs/1704.05179} {{SearchQA}: {A} new q{\&}a
  dataset augmented with context from a search engine}.
\newblock In \emph{Proceedings of the 54th Annual Meeting of the Association
  for Computational Linguistics (ACL)}.

\bibitem[{Fisch et~al.(2019)Fisch, Talmor, Jia, Seo, Choi, and Chen}]{mrqa}
Adam Fisch, Alon Talmor, Robin Jia, Minjoon Seo, Eunsol Choi, and Danqi Chen.
  2019.
\newblock \href {https://doi.org/10.18653/v1/D19-5801} {{MRQA} 2019 shared
  task: Evaluating generalization in reading comprehension}.
\newblock In \emph{Proceedings of 2nd Machine Reading for Reading Comprehension
  (MRQA) Workshop at EMNLP}, pages 1--13.

\bibitem[{Gardner et~al.(2018)Gardner, Grus, Neumann, Tafjord, Dasigi, Liu,
  Peters, Schmitz, and Zettlemoyer}]{allen}
Matt Gardner, Joel Grus, Mark Neumann, Oyvind Tafjord, Pradeep Dasigi,
  Nelson~F. Liu, Matthew Peters, Michael Schmitz, and Luke Zettlemoyer. 2018.
\newblock \href {https://doi.org/10.18653/v1/W18-2501} {{A}llen{NLP}: A deep
  semantic natural language processing platform}.
\newblock In \emph{Proceedings of Workshop for {NLP} Open Source Software
  ({NLP}-{OSS})}, pages 1--6.

\bibitem[{He et~al.(2017)He, Lee, Lewis, and Zettlemoyer}]{hesrl}
Luheng He, Kenton Lee, Mike Lewis, and Luke Zettlemoyer. 2017.
\newblock \href {https://doi.org/10.18653/v1/P17-1044} {Deep semantic role
  labeling: What works and what{'}s next}.
\newblock In \emph{Proceedings of the 55th Annual Meeting of the Association
  for Computational Linguistics (ACL)}, pages 473--483.

\bibitem[{Hudson and Manning(2018)}]{mac}
Drew~A. Hudson and Christopher~D. Manning. 2018.
\newblock Compositional attention networks for machine reasoning.
\newblock In \emph{Proceedings of the 7th International Conference on Learning
  Representations (ICLR)}.

\bibitem[{Joshi et~al.(2019)Joshi, Chen, Liu, Weld, Zettlemoyer, and
  Levy}]{spanbert}
Mandar Joshi, Danqi Chen, Yinhan Liu, Daniel~S. Weld, Luke Zettlemoyer, and
  Omer Levy. 2019.
\newblock Spanbert: Improving pre-training by representing and predicting
  spans.
\newblock In \emph{CoRR}, volume abs/1907.10529.

\bibitem[{Joshi et~al.(2017)Joshi, Choi, Weld, and Zettlemoyer}]{TriviaQA}
Mandar Joshi, Eunsol Choi, Daniel~S. Weld, and Luke Zettlemoyer. 2017.
\newblock {TriviaQA}: {A} large scale distantly supervised challenge dataset
  for reading comprehension.
\newblock In \emph{Proceedings of the 55th Annual Meeting of the Association
  for Computational Linguistics (ACL)}.

\bibitem[{Khashabi et~al.(2018)Khashabi, Khot, Sabharwal, and Roth}]{view}
Daniel Khashabi, Tushar Khot, Ashish Sabharwal, and Dan Roth. 2018.
\newblock {Question Answering as Global Reasoning over Semantic Abstractions}.
\newblock In \emph{{The Thirty-Fourth AAAI Conference on Artificial
  Intelligence (AAAI)}}.

\bibitem[{Kočiský et~al.(2018)Kočiský, Schwarz, Blunsom, Dyer, Hermann,
  Melis, and Grefenstette}]{narrativeQA}
Tomáš Kočiský, Jonathan Schwarz, Phil Blunsom, Chris Dyer, Karl~Moritz
  Hermann, Gábor Melis, and Edward Grefenstette. 2018.
\newblock \href {https://doi.org/10.1162/tacl_a_00023} {{The NarrativeQA
  Reading Comprehension Challenge}}.
\newblock In \emph{Transactions of the Association for Computational
  Linguistics (TACL)}, volume~6, page 317–328.

\bibitem[{Kwiatkowski et~al.(2019)Kwiatkowski, Palomaki, Redfield, Collins,
  Parikh, Alberti, Epstein, Polosukhin, Kelcey, Devlin, Lee, Toutanova, Jones,
  Chang, Dai, Uszkoreit, Le, and Petrov}]{naturalQuestions}
Tom Kwiatkowski, Jennimaria Palomaki, Olivia Redfield, Michael Collins, Ankur
  Parikh, Chris Alberti, Danielle Epstein, Illia Polosukhin, Matthew Kelcey,
  Jacob Devlin, Kenton Lee, Kristina~N. Toutanova, Llion Jones, Ming-Wei Chang,
  Andrew Dai, Jakob Uszkoreit, Quoc Le, and Slav Petrov. 2019.
\newblock \href
  {https://tomkwiat.users.x20web.corp.google.com/papers/natural-questions/main-1455-kwiatkowski.pdf}
  {Natural questions: a benchmark for question answering research}.
\newblock In \emph{Transactions of the Association of Computational Linguistics
  (TACL)}.

\bibitem[{Liu and Perez(2017)}]{gatedmemory}
Fei Liu and Julien Perez. 2017.
\newblock \href {https://www.aclweb.org/anthology/E17-1001} {{Gated End-to-End
  Memory Networks}}.
\newblock In \emph{Proceedings of the 15th Conference of the {E}uropean Chapter
  of the Association for Computational Linguistics (EACL)}, pages 1--10.

\bibitem[{Mihaylov and Frank(2019)}]{discourseaware}
Todor Mihaylov and Anette Frank. 2019.
\newblock \href {https://doi.org/10.18653/v1/D19-1257} {Discourse-aware
  semantic self-attention for narrative reading comprehension}.
\newblock In \emph{Proceedings of the 2019 Conference on Empirical Methods in
  Natural Language Processing and 9th International Joint Conference on Natural
  Language Processing, (EMNLP-IJCNLP)}, pages 2541--2552.

\bibitem[{Narayanan and Harabagiu(2004)}]{semanticstructure}
Srini Narayanan and Sanda Harabagiu. 2004.
\newblock \href {https://www.aclweb.org/anthology/C04-1100} {Question answering
  based on semantic structures}.
\newblock In \emph{Proceedings of the 13th International Conference on
  Computational Linguistics (COLING)}, pages 693--701.

\bibitem[{Palmer et~al.(2005)Palmer, Gildea, and Kingsbury}]{propbank}
Martha Palmer, Daniel Gildea, and Paul Kingsbury. 2005.
\newblock \href {https://doi.org/10.1162/0891201053630264} {The proposition
  bank: An annotated corpus of semantic roles}.
\newblock In \emph{Computational Linguistics (CL)}, pages 71--106.

\bibitem[{Rajpurkar et~al.(2018)Rajpurkar, Jia, and Liang}]{2squad}
Pranav Rajpurkar, Robin Jia, and Percy Liang. 2018.
\newblock \href {https://doi.org/10.18653/v1/P18-2124} {Know what you don{'}t
  know: Unanswerable questions for {SQ}u{AD}}.
\newblock In \emph{Proceedings of the 56th Annual Meeting of the Association
  for Computational Linguistics}, pages 784--789.

\bibitem[{Rajpurkar et~al.(2016)Rajpurkar, Zhang, Lopyrev, and Liang}]{squad1}
Pranav Rajpurkar, Jian Zhang, Konstantin Lopyrev, and Percy Liang. 2016.
\newblock {SQuAD: 100,000+ Questions for Machine Comprehension of Text}.
\newblock In \emph{Proceedings of the 2009 Conference on Empirical Methods in
  Natural Language Processing (EMNLP)}, pages 2383--2392.

\bibitem[{Santoro et~al.(2017)Santoro, Raposo, Barrett, Malinowski, Pascanu,
  Battaglia, and Lillicrap}]{relation}
Adam Santoro, David Raposo, David~G Barrett, Mateusz Malinowski, Razvan
  Pascanu, Peter Battaglia, and Timothy Lillicrap. 2017.
\newblock {A simple neural network module for relational reasoning}.
\newblock In \emph{{Advances in Neural Information Processing Systems 30}},
  pages 4967--4976.

\bibitem[{Shen and Lapata(2007)}]{semanticrole}
Dan Shen and Mirella Lapata. 2007.
\newblock \href {https://www.aclweb.org/anthology/D07-1002} {Using semantic
  roles to improve question answering}.
\newblock In \emph{Proceedings of the 2009 Conference on Empirical Methods in
  Natural Language Processing ({EMNLP}-{C}o{NLL})}, pages 12--21.

\bibitem[{Su et~al.(2019)Su, Xu, Winata, Xu, Kim, Liu, and Fung}]{hltc}
Dan Su, Yan Xu, Genta~Indra Winata, Peng Xu, Hyeondey Kim, Zihan Liu, and
  Pascale Fung. 2019.
\newblock \href {https://doi.org/10.18653/v1/D19-5827} {Generalizing question
  answering system with pre-trained language model fine-tuning}.
\newblock In \emph{Proceedings of the 2nd Workshop on Machine Reading for
  Question Answering (EMNLP)}, pages 203--211.

\bibitem[{Sun et~al.(2018)Sun, Cheng, and Qu}]{sun-reading}
Yawei Sun, Gong Cheng, and Yuzhong Qu. 2018.
\newblock \href {https://www.aclweb.org/anthology/C18-1069} {{Reading
  Comprehension with Graph-based Temporal-Casual Reasoning}}.
\newblock In \emph{Proceedings of the 27th International Conference on
  Computational Linguistics (COLING)}, pages 806--817.

\bibitem[{Sun et~al.(2019)Sun, Wang, Li, Feng, Chen, Zhang, Tian, Zhu, Tian,
  and Wu}]{baidu_ernie}
Yu~Sun, Shuohuan Wang, Yukun Li, Shikun Feng, Xuyi Chen, Han Zhang, Xin Tian,
  Danxiang Zhu, Hao Tian, and Hua Wu. 2019.
\newblock \href {http://arxiv.org/abs/1904.09223} {{ERNIE:} enhanced
  representation through knowledge integration}.
\newblock In \emph{CoRR}, volume abs/1904.09223.

\bibitem[{Takahashi et~al.(2019)Takahashi, Taniguchi, Taniguchi, and
  Ohkuma}]{cler}
Takumi Takahashi, Motoki Taniguchi, Tomoki Taniguchi, and Tomoko Ohkuma. 2019.
\newblock \href {https://www.aclweb.org/anthology/D19-5824} {{CLER}: Cross-task
  learning with expert representation to generalize reading and understanding}.
\newblock In \emph{Proceedings of the 2nd Workshop on Machine Reading for
  Question Answering}, pages 183--190.

\bibitem[{Trischler et~al.(2017)Trischler, Wang, Yuan, Harris, Sordoni,
  Bachman, and Suleman}]{newsQA}
Adam Trischler, Tong Wang, Xingdi Yuan, Justin Harris, Alessandro Sordoni,
  Philip Bachman, and Kaheer Suleman. 2017.
\newblock {NewsQA: A Machine Comprehension Dataset}.
\newblock In \emph{Proceedings of the 2nd Workshop on Representation Learning
  for NLP}, pages 191--200.

\bibitem[{Wang et~al.(2019)Wang, Singh, Michael, Hill, Levy, and Bowman}]{glue}
Alex Wang, Amanpreet Singh, Julian Michael, Felix Hill, Omer Levy, and
  Samuel~R. Bowman. 2019.
\newblock \href {http://arxiv.org/abs/1804.07461} {{GLUE:} {A} multi-task
  benchmark and analysis platform for natural language understanding}.
\newblock In \emph{The International Conference on Learning Representations
  (ICLR)}.

\bibitem[{Weston et~al.(2014)Weston, Chopra, and Bordes}]{memory}
Jason Weston, Sumit Chopra, and Antoine Bordes. 2014.
\newblock {Memory Networks}.
\newblock In \emph{Proceedings of the 3rd International Conference on Learning
  Representations (ICLR)}.

\bibitem[{Williams et~al.(2018)Williams, Nangia, and Bowman}]{mnli}
Adina Williams, Nikita Nangia, and Samuel Bowman. 2018.
\newblock \href {https://doi.org/10.18653/v1/N18-1101} {A broad-coverage
  challenge corpus for sentence understanding through inference}.
\newblock In \emph{Proceedings of the 2018 Conference of the North {A}merican
  Chapter of the Association for Computational Linguistics: Human Language
  Technologies (NAACL)}, pages 1112--1122.

\bibitem[{Yih et~al.(2013)Yih, Chang, Meek, and Pastusiak}]{lexical}
Wen-tau Yih, Ming-Wei Chang, Christopher Meek, and Andrzej Pastusiak. 2013.
\newblock \href {https://www.aclweb.org/anthology/P13-1171} {{Question
  Answering Using Enhanced Lexical Semantic Models}}.
\newblock In \emph{Proceedings of the 44th Annual Meeting of the Association
  for Computational Linguistics (ACL)}, pages 1744--1753.

\bibitem[{Yu et~al.(2019)Yu, Zha, and Yin}]{inferentialrace}
Jianxing Yu, Zhengjun Zha, and Jian Yin. 2019.
\newblock \href {https://doi.org/10.18653/v1/P19-1217} {Inferential machine
  comprehension: Answering questions by recursively deducing the evidence chain
  from text}.
\newblock In \emph{Proceedings of the 57th Annual Meeting of the Association
  for Computational Linguistics (ACL)}, pages 2241--2251.

\bibitem[{Zhang et~al.(2019)Zhang, Han, Liu, Jiang, Sun, and Liu}]{infoernie}
Zhengyan Zhang, Xu~Han, Zhiyuan Liu, Xin Jiang, Maosong Sun, and Qun Liu. 2019.
\newblock \href {https://doi.org/10.18653/v1/P19-1139} {{ERNIE}: Enhanced
  language representation with informative entities}.
\newblock In \emph{Proceedings of the 50th Annual Meeting of the Association
  for Computational Linguistics (ACL)}, pages 1441--1451.

\bibitem[{Zhang et~al.(2018)Zhang, Wu, Li, and Zhao}]{paclic}
Zhuosheng Zhang, Yuwei Wu, Zuchao Li, and Hai Zhao. 2018.
\newblock \href {https://arxiv.org/abs/1809.02794} {{Explicit Contextual
  Semantics for Text Comprehension}}.
\newblock In \emph{Proceedings of the 33nd Pacific Asia Conference on Language,
  Information and Computation (PACLIC 33)}.

\bibitem[{Zhang et~al.(2020)Zhang, Wu, Zhao, Li, Zhang, Zhou, and
  Zhou}]{sembert}
Zhuosheng Zhang, Yuwei Wu, Hai Zhao, Zuchao Li, Shuailiang Zhang, Xi~Zhou, and
  Xiang Zhou. 2020.
\newblock {Semantics-aware BERT for Language Understanding}.
\newblock In \emph{{The Thirty-Fourth AAAI Conference on Artificial
  Intelligence (AAAI)}}.

\end{thebibliography}
\bibliographystyle{acl_natbib}
\appendix
\end{document}